%% file: arxivmain.tex
\documentclass{article}
\usepackage{subfiles}

\usepackage[nonatbib,preprint]{neurips_2023}

\usepackage[utf8]{inputenc} 
\usepackage[T1]{fontenc}    
\usepackage{hyperref}       
\usepackage{url}            
\usepackage{booktabs}       
\usepackage{amsfonts}       
\usepackage{amsmath}
\usepackage{nicefrac}       
\usepackage{microtype}      
\usepackage{xcolor}         
\usepackage{graphicx}  
\usepackage{wrapfig}
\usepackage{multirow}
\usepackage{sidecap}
\sidecaptionvpos{figure}{c}

\newcommand{\methodName}{\texttt{ImageSelect }}
\newcommand{\methodNameNS}{\texttt{ImageSelect}}
\newcommand{\myparagraph}[1]{\vspace{1pt}\noindent{\bf #1}}
\usepackage{amsmath}
\DeclareMathOperator*{\argmax}{arg\,max}

\title{If at First You Don't Succeed, Try, Try Again:\\ Faithful Diffusion-based Text-to-Image Generation\\ by Selection}

\author{
Shyamgopal Karthik$^{*, 1}$, Karsten Roth$^{*, 1}$, Massimiliano Mancini$^2$, Zeynep Akata$^{1,3}$\\
$^1$University of Tübingen $^2$University of Trento $^3$MPI for Intelligent Systems\\
$^*$equal contribution
}

\begin{document}

\maketitle

\begin{abstract}
Despite their impressive capabilities, diffusion-based text-to-image (T2I) models can lack faithfulness to the text prompt, where generated images may not contain all the mentioned objects, attributes or relations. 
To alleviate these issues, recent works proposed post-hoc methods to improve model faithfulness without costly retraining, by modifying how the model utilizes the input prompt.
In this work, we take a step back and show that large T2I diffusion models \textit{are more faithful than usually assumed}, and can generate images faithful to even complex prompts without the need to manipulate the generative process.
Based on that, we show how faithfulness can be simply treated as a candidate selection problem instead, and introduce a straightforward pipeline that generates candidate images for a text prompt and picks the best one according to an automatic scoring system that can leverage already existing T2I evaluation metrics. 
Quantitative comparisons alongside user studies on diverse benchmarks show consistently improved faithfulness over post-hoc enhancement methods, with comparable or lower computational cost. Code is available at \url{https://github.com/ExplainableML/ImageSelect}.
\end{abstract}

\section{Introduction}
Text-to-Image (T2I) Generation \cite{mansimov2015generating,reed2016generative,zhang2017stack} has seen drastic progress in recent times with the advent of modern generative models. Starting from GAN-based \cite{goodfellow2014gan} approaches \cite{reed2016generative,zhang2017stack}, this process was supercharged and popularized with the release of Stable Diffusion~\cite{rombach22ldm} and other large-scale pretrained generative models~\cite{bommasani2021opportunities,sauer2023stylegant,dall-e2,makeascene,parti,kang2023scaling}. However, even these large models appear to exhibit shortcomings, particularly when it comes to faithfully generating the input prompt, failing to correctly reflect attributes, counts, semantic object relations or even entire objects~\cite{liu2022compositional,feng2023structured,chefer2023attendandexcite}. 
Consequently, recent works such as Composable Diffusion~\cite{liu2022compositional}, Structure Diffusion~\cite{feng2023structured}, Space-Time Attention~\cite{wu2023spacetime} or Attend-and-Excite~\cite{chefer2023attendandexcite} propose to improve faithfulness in these baseline models by modifying the inference procedure. While resulting in a more expensive generation process (e.g. Attend-and-Excite \cite{chefer2023attendandexcite} being around six times slower, and \cite{wu2023spacetime} over a hundred times), qualitative demonstrations showcase superior faithfulness compared to the baselines. However, these methods are often tailored to special prompt types. Paired with the mostly qualitative support, it remains unclear if they can work in general-purpose settings with a larger and more diverse set of prompts.

As such, in this work, we take a step back and investigate how unfaithful these diffusion models really are.
Upon closer inspection, we observe that the faithfulness of Stable Diffusion is affected heavily by the random seed that determines the initial latent noise, suggesting that within the explorable latent space, faithful image generations are possible (c.f. for example image candidates in Fig.~\ref{fig:first_page}).
Motivated by this observation, we thus propose to improve the faithfulness in diffusion models not through an explicit change in the baseline model, but instead by simply querying it multiple times and finding ways to automatically select the most suitable output. We denote this simple pipeline as \texttt{ImageSelect}.
We utilize metrics from recently proposed text-to-image faithfulness benchmarks, TIFA~\cite{hu2023tifa} and ImageReward~\cite{xu2023imagereward}, to evaluate the faithfulness of our image generation. TIFA simplifies the text-to-image matching process into a set of Visual Question Answering tasks, which can be more easily solved with existing pretrained models than the complex input prompts used in direct matching. ImageReward proposes a matching model trained on human preferences, where candidates assign preference scores to generated images. In both cases, the matching qualities are significantly better than those of previous approaches that use global image-text matching with a vision-language model, such as CLIPScore~\cite{hessel2021clipscore} or CLIP-R-Precision~\cite{compt2i}. Our results with these metrics provide evidence that candidate selection can improve faithfulness, and  improvements in faithfulness measures can directly translate to better generation faithfulness using \texttt{ImageSelect}.

\input{figures/first_page}

To understand the efficacy of \texttt{ImageSelect}, we first study each selection mechanism against all reference methods evaluated with opposing metrics - TIFA as the selection mechanism evaluated on the ImageReward metric, and vice versa. To ensure sufficient generality of our results, we generate a diverse collection of over 1000 prompts, \texttt{diverse-1k}, aggregated from multiple datasets (HRS~\cite{hrs}, TIFA~\cite{hu2023tifa}/MSCOCO~\cite{mscoco}, Structure Diffusion~\cite{feng2023structured}), spanning different textual aspects such as counting, spatial relations and attribute binding. Doing so also mitigates overfitting to a particular prompt generation approach from a specific dataset.
Results on \texttt{diverse-1k} in both cases indicate significant performance improvements against reference methods, with gains in faithfulness through automatic candidate selection consistently higher than that even achieved by changed model version generations (going for example from Stable Diffusion 1.4 to 2.1). This improvement in faithfulness holds even when investigating faithfulness for specific prompt types. 
In addition, we perform an extensive human evaluation in which \texttt{ImageSelect} is compared against baseline methods on human-evaluated faithfulness. Results produced by over 5000 image comparisons covering 68 voluntary participants strongly support our observations made on the quantitative tests, with \texttt{ImageSelect} outputs preferred in parts over three times as often as baseline method outputs. 
The results showcase a simple, but large step forward for text-to-image faithfulness, and highlight our insights as a crucial sanity check for future work tackling the task of post-hoc enhancement of text-to-image generation.

To summarize, we make the following contributions:
(1) We highlight that, given a prompt, the faithfulness (and quality) of images generated by diffusion-based text-to-image generative approaches varies significantly across multiple generations with different seeds.
(2) From this insight, we propose \texttt{ImageSelect}, a simple pipeline which generates multiple candidate images and selects the most faithful one via an automatic scoring mechanism.
(3) Quantitative studies and extensive user studies on diverse benchmarks show that \texttt{ImageSelect} significantly outperforms existing methods in text-to-image faithfulness while matching or even improving their inference speeds.

\section{Related Work}
\myparagraph{Faithful Text-to-Image Generation.} T2I generation was first introduced with GAN~\cite{goodfellow2014gan} models generalizing to unseen concepts~\cite{reed2016generative,reed2016learning,stackgan}. Later works explored other generative architectures such as VQ-VAE/VQ-GANs~\cite{esser2021taming,dall-e,vqgan,makeascene,lee2022autoregressive,gu2022vector,alaniz2022semantic} and diffusion models~\cite{ogddpm,ho2020denoising, dhariwal2021diffusion,glide,rombach22ldm,dall-e2,imagen}. The latter dominate the current state-of-the-art, with text conditioning coming from either a language~\cite{t5} or a vision-language~\cite{clip} model. However, even these advanced methods struggle to capture detailed prompt semantics, such as composing arbitrary concepts, counting~\cite{paiss2023teaching}, spelling~\cite{spelling}, and handling biases~\cite{bias1,bias2}.
Recent works address these shortcomings post-hoc by changing the latent diffusion process in models s.a. Stable Diffusion \cite{rombach22ldm} or DALL-E 2 \cite{dall-e2}. Composable Diffusion \cite{liu2022compositional} handles conjunction and negation operations by recomposing diffusion outputs at every timestep. Structure Diffusion \cite{feng2023structured} performs multi-guidance via CLIP~\cite{clip} text embeddings of different noun phrases in a prompt. Attend-and-Excite \cite{chefer2023attendandexcite} optimizes cross-attention maps~\cite{hertz2022prompt}, ensuring they attend to manually selected prompt parts. Space-Time Attention~\cite{wu2023spacetime} improves faithfulness with a separate layout predictor and temporal attention control.
Unlike these approaches, we found that T2I diffusion models s.a. Stable Diffusion already exhibit a large degree of faithfulness that a simple and automatic candidate selection process can capture without altering the generative process.

\myparagraph{Evaluating Image-Text Alignment.} Large vision-language models~\cite{clip,ilharco2021openclip,sun2023evaclip} offer direct tools to evaluate and leverage image-text alignment (e.g. \cite{hessel2021clipscore,roth2022lang,yang2022lang,elBanani2023lang}), but lack compositional understanding~\cite{vlmbow}.
Other approaches~\cite{bleu,anderson2016spice,banerjee2005meteor,cider} propose to caption the generated image and measure the textual similarity between the prompt and caption. However, these metrics are not well correlated with human preferences~\cite{compt2i,hu2023tifa,humaneval}, and may
miss fine-grained details of the prompt. Inspired by the success of reinforcement learning from human feedback~\cite{ogrlfh,openairlhf,stiennon2020learning}, several works~\cite{xu2023imagereward,lee2023aligning,hps} trained models to predict human preferences instead. However, this requires expensive annotations, while not disentangling preferences regarding the quality of the generation and faithfulness to the prompt. Instead, TIFA~\cite{hu2023tifa} measures faithfulness by answering questions about the prompt using a VQA model (s.a. BLIP~\cite{blip,blip2}), producing a fine-grained and interpretable rating.
These metrics are part of ongoing efforts to provide quantitative benchmarks for T2I models, s.a. MS-COCO~\cite{mscoco,cococaptions}, Comp-T2i~\cite{compt2i}, DALL-E-Eval~\cite{dalleeval}, HRS~\cite{hrs}, VSR~\cite{vsr}, TIFA~\cite{hu2023tifa}, CC~\cite{feng2023structured}, ABC~\cite{feng2023structured}, PaintSkill~\cite{dalleeval}, DrawBench~\cite{imagen}, PartiPrompts~\cite{parti} or VISOR~\cite{gokhale2022benchmarking}. 
To ensure the generality of our results beyond the prompt generation process of a single dataset, we also leverage an aggregate prompt collection using TIFA, MS-COCO, HRS, and Structure Diffusion to test general-purpose T2I faithfulness across a wide range of categories. 

\section{Achieving Faithfulness through Selection}
\input{figures/arch}

We first provide an overview of Latent Diffusion Models and a motivation for faithfulness through candidate selection. From these findings, we describe measures for text-to-image alignment and how they can be used to improve T2I faithfulness via selection. Finally, we provide details for our diverse benchmark, \texttt{diverse-1k}, which we use in the experiments to validate our findings.

\subsection{Background: Latent Diffusion Models}
Latent Diffusion Models (LDMs)~\cite{rombach22ldm} extend Denoising Diffusion Probabilistic Models (DDPM)~\cite{ho2020denoising} into the latent space of pretrained encoder-decoder models s.a. VAEs \cite{kingma2014vae}, where the compression allows for improved scalability. 
Unlike generic DDPMs which model the generation of an image $x_0$ as an iterative denoising process with $T$ steps starting from noise $x_T$ (sampled from a Normal prior), LDMs deploy the denoising process over spatial latents $z_T \rightarrow z_0$ of the pretrained model. Starting from $z_T$, these LDMs (often parametrized as a UNet \cite{unet} with parameters $\theta$) provide a perturbation $\epsilon_\theta(z_t, t)$ for every timestep $t\in[1, ..., T]$, which is subtracted from $z_t$ to generate subsequent latents 
\begin{equation}
    z_{t-1} = z_t - \epsilon_\theta(z_t, t) + \mathcal{N}(0, \sigma^2_tI)    
\end{equation}
with learned covariances $\sigma^2_tI$. When $z_0$ is reached, the decoder projects the latent back into the image space. The favorable scaling properties of operating in latent spaces allow LDMs to produce large-scale pretrained, high-quality generative models such as Stable Diffusion \cite{rombach22ldm}. 
Additional text-conditioning can then be performed during the denoising process. For Stable Diffusion, this condition is simply a text embedding produced by CLIP \cite{clip}, $c(y)$, corresponding to associated prompts $y$. By extending the standard UNet with cross-attention layers (e.g. \cite{hertz2022prompt,chefer2023attendandexcite,feng2023structured,chefer2021generic}) to connect these embeddings with the latent features, the text-conditioned LDM can then simply be trained in the same manner as standard LDMs.
While these LDMs can generate high-quality images when trained at scale, recent works \cite{liu2022compositional,feng2023structured,chefer2023attendandexcite,wu2023spacetime} strongly emphasize that they lack faithfulness to the text prompt, as shown in a qualitative fashion on specific input prompts and seeds.

\subsection{\texttt{ImageSelect}: Faithfulness through Selection}
Indeed, our first qualitative study on various prompts over multiple seeds using vanilla Stable Diffusion indicates that faithful images \textit{can be} generated, but are simply hidden behind a suitable selection of the starting latent noise (see Fig.~\ref{fig:first_page}).
Based on this insight, we thus introduce a simple, efficient and effective mechanism to provide more faithful outputs for a given prompt by simply looking at candidates from multiple seeds and automatically selecting the most suitable image.

\myparagraph{Measuring Faithfulness in Text-to-Image Alignment.}
For our automatic selection, we show that one can simply leverage already existing advanced T2I evaluation methods. As \textit{proof-of-concept}, we simply select two - TIFA and ImageReward - which we explain in the following in more detail.

TIFA Scores~\cite{hu2023tifa} evaluate T2I alignment using the auxiliary task of Visual-Question Answering (VQA)~\cite{vqa}. Specifically, given a text prompt $y$, and a generated image $I$, a Large Language Model (LLM) such as GPT3.5~\cite{gpt3} is used to generate
question-answer pairs $\mathcal{Q}(y) := \{(Q_i, A_i)\}_i$ related to the prompt or caption $y$ \cite{changpinyo2022all}. An off-the-shelf VQA model $\Psi_\text{VQA}$ such as BLIP~\cite{blip,blip2} or mPLUG~\cite{mplug} is then used to answer these generated questions using the generated image $I$, providing respective answers $A^\text{VQA}_i$ for given questions $Q_i$. Doing so breaks down the matching process into many easier-to-solve, small-scale matching problems. The resulting faithfulness score $\mathcal{F}$ of the generated image $I$ is simply defined as the ratio of questions that the VQA model answered correctly, 
\begin{equation}\label{eq:tifa}
    \mathcal{F}_\text{TIFA}(I,y) = \frac{1}{|\mathcal{Q}(y)|}\sum_{(Q_i, A_i)\sim\mathcal{Q}(y)} \mathbb{I}\left[\Psi_\text{VQA}(I, Q_i) = A_i\right].
\end{equation}
where $\mathbb{I}\left[\Psi_\text{VQA}(I, Q_i) = A_i\right]$ is 1 if the answer is correct. This evaluation strategy has the benefits of being interpretable, fine-grained, and avoiding any manual annotations for text-image alignment.

ImageReward Scores~\cite{xu2023imagereward} are produced from a completely different direction, following more closely the trend of just end-to-end training on suitable data. In particular, \cite{xu2023imagereward} simply train a Multi-Layer Perception (MLP) on top of image and text features produced by BLIP to regress 137k expert human preference scores on image-text pairs, with higher scores denoting higher levels of faithfulness. The resulting rating model $\Psi_\text{ImageReward}$, while not normalized, is well-correlated with human ratings even on samples outside the training dataset, and gives the faithfulness score simply as 
\begin{equation}\label{eq:imgreward}
    \mathcal{F}_\text{ImageReward}(I,y) = \Psi_\text{ImageReward}(I,y).
\end{equation}

\myparagraph{Faithfulness through Selection.} Both TIFA and ImageReward are only utilized as a benchmarking mechanism to evaluate current and future T2I methods on faithfulness. Instead, we showcase that these metrics can be easily utilized to supercharge the faithfulness of existing models without any additional retraining, by simply re-using them in a contrastive framework as a candidate selection metric. In particular, given a budget of $N$ initialization starting points and a text prompt $y$, our associated generated output image $I$ is thus simply given as 
\begin{equation}\label{eq:imageselect}
    I_\texttt{ImageSelect}(y) = \argmax_{n\in N} \mathcal{F}_\texttt{ImageSelect}\left(\mathcal{D}(\epsilon_\theta(\epsilon_n, T, y)),y\right)
\end{equation}
where $\epsilon_\theta$ denotes the text-conditioned denoising diffusion model in the latent space of the encoder-decoder model with decoder $\mathcal{D}$, total number of denoising iterations $T$, and initial latent noise $\epsilon_n\sim\mathcal{N}(0,1)$ sampled anew for each $n$. We note that we use \texttt{ImageSelect} to refer to the use of any faithfulness measure s.a. $\mathcal{F}_\text{TIFA}$, $\mathcal{F}_\text{ImageReward}$, and highlight that this can be extended to any other scoring mechanism or combinations thereof. 
For a given selection method, we denote the respective \texttt{ImageSelect} operation as \texttt{TIFASelect} or \texttt{RewardSelect}.

\input{tables/diverse_prompts}
\subsection{The Diverse Prompts Dataset}
While multiple benchmarks have recently been proposed to study text-to-image faithfulness, most benchmarks introduce their unique sets of prompts. These are grouped under different fine- or coarse-grained categories like \textit{shape}, \textit{attribute} or \textit{color} in TIFA, which are shared in e.g. HRS \cite{hrs}, or more general prompt types such as \textit{emotions} or \textit{long prompts} specifically introduced in HRS. 
To ensure that our results are as representative as possible and do not overfit to a particular type of prompt generation mechanism introduced in a benchmark, we aggregate prompts from HRS, TIFA (containing also captions from MS-COCO), and prompts utilized in \cite{feng2023structured}. 
Given the higher diversity and count of prompts in HRS and TIFA, we oversample from both. For HRS, we cover each sub-category. We avoid duplicates or semantic equivalents by first filtering based on language similarity (using a CLIP text encoder) before manual removal. 
We plan to release the prompt collection to aid future research on faithful text-to-image generation.

\section{Experiments}
\myparagraph{Implementation Details.} We take off-the-shelf Stable Diffusion 1.4 and 2.1 and evaluate them on the TIFAv1.0~\cite{hu2023tifa} benchmark - consisting of prompts from MS-COCO and other sources that benchmark T2I generation for more creative tasks - and our \texttt{diverse-1k} prompts list. We consider the Structure Diffusion (StrD)~\cite{feng2023structured} \& Composable Diffusion (CD)~\cite{liu2022compositional} (both available only with Stable Diffusion 1.4) and the Attend-and-Excite (A\&E)~\cite{chefer2023attendandexcite} methods as our baselines. While StrD can be applied directly, CD requires us to split the prompts and join them together using the ``AND'' operator. 

\myparagraph{Extending Attend \& Excite for automatic usage.} A\&E requires a user to manually select tokens the model should attend to. We modify this to work automatically by selecting categories from MS-COCO, as well as utilizing NLTK~\cite{loper2002nltk} to determine nouns which cannot be treated as either a verb or adjective. For any prompt for which the above protocol provides no target tokens, we continuously relax the constraints over the nouns. In limit cases where nothing suitable is selected, A\&E defaults back to the original Stable Diffusion it extends. We denote A\&E equipped with this formalism as \textit{Attend-and-Excite++} (A\&E++). We find that on normal prompts or those qualitatively studied in the original paper \cite{chefer2023attendandexcite}, our protocol comes very close to the generations reported in \cite{chefer2023attendandexcite}.

\subsection{Quantitative comparison between Stable Diffusion variants}

\input{figures/metric_comp}

\myparagraph{Faithfulness on \texttt{diverse-1k}.} We begin by evaluating the faithfulness of baselines on top of Stable Diffusion Version 1.4 (SD1.4) and Version 2.1 (SD2.1, where possible) on \texttt{diverse-1k}, which we evaluate using both TIFA (eq.~\ref{eq:tifa}) and ImageReward score (eq.~\ref{eq:imgreward}). 
We use \texttt{RewardSelect} for TIFA scores, and vice versa \texttt{TIFASelect} for the ImageReward score evaluation, over a pool of 10 randomly generated images per prompt to evaluate the quantitative impact of \texttt{ImageSelect}.
Results in Fig.~\ref{fig:metrics_comp} highlight a \textbf{clear} increase in faithfulness of \texttt{ImageSelect} over all baseline methods across both evaluation metrics. 
We also find that across diverse, non-cherry-picked prompts, both Composable and Structure Diffusion can actually have an overall detrimental effect, 
\input{figures/detail_comp}
with standard SD1.4 scoring $71.6\%$ on TIFA and $-0.22$ on ImageReward, and Structure Diffusion only $70.6\%$ on TIFA and $-0.35$ on ImageReward.
For Composable Diffusion, performance also falls below on ImageReward ($-0.35$). 
On the opposite end, we find our extension of \cite{chefer2023attendandexcite}, Attend-and-Excite++, to offer faithfulness benefits (e.g. $75.2\%$ TIFA score) across SD1.4 and SD2.1.  
However, this change in performance is overshadowed by \texttt{ImageSelect}, which e.g. on SD1.4 achieves an impressive $80.4\%$ - over $4pp$ \textit{higher than the change from SD1.4 to SD2.1} gives in terms of text-to-image faithfulness. 
This fact is only exacerbated on the ImageReward score ($-0.22$ SD1.4, $0.18$ SD2.1 and $0.32$ for \texttt{TIFASelect}). 
Together, these results provide a first clear quantitative indicator that suitable candidate selection can have a much higher impact on faithfulness than current explicit changes to the generative process. 
For completeness, we test simple CLIPScore selection (in the same fashion as Eq.\ref{eq:imageselect}) against \texttt{RewardSelect} on TIFA ($72.9\%$ versus $80.8\%$ and $71.6\%$ for SD V1.4), and against \texttt{TIFASelect} on ImageReward ($-0.129$ vs $0.316$ and $-0.22$ for standard Stable Diffusion V1.4). As can be seen, while faithfulness over the Stable Diffusion baseline is increased, the overall performance falls short compared to more suitable selection mechanisms. We believe these insights hint towards the potential impact of further research into selection approaches to improve faithfulness.

\myparagraph{Breakdown by Categories.} We repeat our previous experiments on the original TIFAv1.0 benchmark \cite{hu2023tifa} (where parts were integrated into \texttt{diverse-1k}), as the benchmark offers easy category-level grouping such as ``counting'', ``spatial (relations)'', ``shape'' etc. While \texttt{diverse-1k} also offers subset breakdowns (c.f. Table~\ref{tab:diverse_prompts}), the grouping in TIFAv1.0 provides a simple, straightforward attribute-style separation. For all methods and \texttt{RewardSelect} on SD1.4, we showcase results in Fig.~\ref{fig:detail_comp}.
When breaking down the overall improvement in faithfulness into respective categories, the benefits of \texttt{ImageSelect} become even clearer. \texttt{ImageSelect} improves over every baseline across every single category, with especially significant changes in categories such as ``counting'' (over $10pp$) - a well-known shortcoming of T2I diffusion models~\cite{paiss2023teaching}. While not a complete remedy, the change in performance is remarkable. 
Similarly, we see other scenarios such as ``spatial (relations)'' or ``object (inclusion)'' improving from $0.71$ to $0.78$ and $0.77$ to $0.85$, respectively. 
Again, it is important to highlight that these improvements are not a result of potential overfitting to the evaluation metric, as the scoring approaches are entirely different (VQA versus modeling human preferences).

\input{tables/ground_truth_comp}

\myparagraph{Comparison to Ground Truth Faithfulness.} To provide a better reference for the quantitative change in performance, 
we also evaluate on the MS-COCO captions used in \cite{hu2023tifa}, for which ground truth images exist. Using \texttt{RewardSelect} and the TIFAScore for evaluation, we report results in Tab.~\ref{tab:ground_truth_comp}. 
While clearly outperforming baseline methods, we also see \texttt{RewardSelect} matching ground truth TIFA faithfulness scores of true MS-COCO image-caption pairs ($89.85\%$ versus $89.09\%$). 
While attributable to increases in measurable faithfulness through \texttt{ImageSelect}, it is important to note both the noise in ground truth captions on MS-COCO \cite{mscoco} and a focus on a particular prompt-style (descriptive natural image captions - hence also our use of \texttt{diverse-1k} for most of this work). Still, these ground truth scores provide strong support for the benefits of candidate selection as a means to increase overall faithfulness.

\input{figures/num_imgs}

\myparagraph{Relation between Faithfulness and Number of Candidate Images.} We further visualize the relation between text-to-image faithfulness and the number of candidate images taken into consideration in Fig.~\ref{fig:seed_comp}, as measured by the ImageReward score on \texttt{diverse-1k}. Our experiments show a drastic improvement with already two candidates, raising the faithfulness of SD1.4 to that of SD2.1. Going further, we find monotonic improvements, but with diminishing returns becoming more evident for larger candidate counts. This also means that a small number of candidate images (e.g. 4) is already sufficient to beat all baselines. We highlight that this is not caused by any single seed being more effective~\cite{picard2021torch}, as we find all seeds to behave similarly ($77.9\%$ to $78.5\%$ for 10 seeds on TIFAv1.0), \textit{but rather the per-prompt candidate selection}.

\myparagraph{Computational Efficiency.} While Stable Diffusion takes 5 seconds to generate a single image (NVIDIA 2080Ti), Attend-and-Excite requires 30 with double the memory requirements. Other recent methods such as Space-Time-Attention~\cite{wu2023spacetime} can require nearly five times the VRAM and over 10 minutes. Thus even from a computational perspective, there is a clear benefit of leveraging simple candidate selection through \texttt{ImageSelect}, and generating as many candidates as possible within a computational budget. Finally, the process of producing respective images for a prompt is parallelizable, and directly benefits from extended GPU counts even on a single-prompt level.

\subsection{User Study}

\input{figures/human_study}

Since quantitative metrics alone can be inadequate for tasks which have subjective choices such as image generation, we expand our quantitative studies with extensive human evaluations. 
For every \texttt{diverse-1k} prompt, we generate images using all baselines (Composable Diffusion \cite{liu2022compositional}, Structure Diffusion \cite{feng2023structured} and Attend-and-Excite++) as well as \texttt{RewardSelect} and \texttt{TIFASelect} on SD1.4. For all \texttt{ImageSelect} variants and Attend-and-Excite++, we also utilize  SD2.1.
%
\input{tables/rel_preferences}
%
Using the generated images, we set up a comparative study following the layout shown in supplementary. Voluntary users interact with the study through a webpage, and are tasked to select the most faithful generation between the output of either a baseline method or an \texttt{ImageSelect} variant. We ensure that the underlying Stable Diffusion model is shared, and the relative positioning on the interface is randomly shuffled for each selection. 
Baseline and \texttt{ImageSelect} method are sampled anew after each choice.
In total, we collect 5093 human preference selections, distributed over 68 unique users and each comparative study. The number of selections performed for a comparative study is between 456 and 538. Results are shown in Fig.~\ref{fig:human_study}, where we also compare \texttt{RewardSelect} and \texttt{TIFASelect} directly.

Looking at the results, we find a clear preference in faithfulness for images generated by \texttt{ImageSelect}, particularly \texttt{RewardSelect}. 
Indeed, when looking at the relative improvements w.r.t. each baseline in Table~\ref{tab:user_study}, we find \texttt{ImageSelect} to be chosen in parts twice (e.g. $+126.3\%$ for \texttt{TIFASelect} vs Comp. Diffusion on SD1.4) or even three times more often (e.g. $+207.9\%$ on \texttt{RewardSelect} vs. Structure Diffusion on SD1.4). Even against our adaptation of \cite{chefer2023attendandexcite} (Attend-and-Excite++) and on the improved Stable Diffusion V2.1, \texttt{RewardSelect} still has a $84.4\%$ higher chance to be chosen as more faithful. 
In general, we found \texttt{RewardSelect} to be better aligned with human insights on text-to-image faithfulness, and better suited as a candidate selector. 
This is further supported when looking at the direct comparisons with \texttt{TIFASelect} in Fig.~\ref{fig:human_study}i-j, and Tab~\ref{tab:user_study}, where \texttt{RewardSelect} is preferred with a $53.6\%$ higher chance on SD V1.4 and $46.5\%$ on SD V2.1. This indicates that a model trained to mimic human preferences might work better as a selection metric than one that looks for faithfulness as a numerical metric, weighing every semantic aspect equally.

Regardless of the variations in \texttt{ImageSelect}, our user study provides compelling evidence that automatic candidate selection is a highly promising approach for post-hoc text-to-image faithfulness in large-scale pretrained text-to-image diffusion models, especially when compared to existing approaches that explicitly adapt the generative process in a costly manner. We intend to publicly release all user preferences collected during the study to facilitate further exploration in this direction.

\input{figures/more_examples}
\input{figures/failure_cases}

\subsection{Qualitative Examples and Limitations}\label{subsec:qualitatives}

We also show additional qualitative examples to illustrate the successes of \methodName in Fig.~\ref{fig:more_examples}, which captures both simple and complex prompts well, particularly compared to other methods that struggle with the issues of catastrophic neglect~\cite{chefer2023attendandexcite}, attribute binding~\cite{feng2023structured}, and incorrect spatial arrangement. For instance, \texttt{\methodNameNS} is able to capture the objects and spatial relations in prompts like \texttt{``three small yellow boxes on a large blue box''} or \texttt{``Two men in yellow jackets near water and a black plane.''}, while also faithfully rendering creative prompts like \texttt{``an oil painting of a cat playing checkers.''}. Other methods perform worse in comparison, often missing objects entirely or generating objects with an incorrect spatial arrangement or false association of attributes (c.f. \texttt{``A green chair and a red horse''}).

\textbf{Limitations.} We illustrate failures in Fig.~\ref{fig:failure_cases}. While \methodName significantly improves faithfulness, it can still struggle with challenges inherent to the underlying model such as rendering text, exact spatial relations, counting or very long prompts. However, due to its applicability to any T2I model, these shortcomings can be addressed by jointly tackling fundamental issues in vision-language models~\cite{vlmbow} and leveraging orthogonal extensions such as e.g. \cite{spelling} for character generation.

\section{Conclusion}
In this work, we both highlight and leverage the dependence of faithfulness on initial latent noises in diffusion-based text-to-image models to introduce \texttt{\methodNameNS}. By viewing the problem of post-hoc faithfulness improvements as a candidate selection problem, we propose a simple pipeline, in which an automatic scoring system selects the most suitable candidate out of multiple model queries. In doing so, we are able to significantly improve faithfulness, particularly when compared to recent approaches adapting the diffusion process directly. 
We validate the success of \texttt{\methodNameNS} with quantitative experiments and user studies on diverse test benchmarks, showcasing significant gains in faithfulness. 
Overall, we hope that our work serves as a useful practical tool and an important reference point for future work on post-hoc enhancement of text-to-image generation.

\section*{Acknowledgements}
This work was supported by DFG project number 276693517, by BMBF FKZ: 01IS18039A, by the ERC (853489 - DEXIM), by EXC number 2064/1 – project number 390727645, and by the MUR PNRR project FAIR - Future AI Research (PE00000013) funded by the NextGenerationEU.
Shyamgopal Karthik and Karsten Roth thank the International Max
Planck Research School for Intelligent Systems (IMPRS-IS) for support. Karsten Roth would also like to thank the European Laboratory for Learning and Intelligent Systems (ELLIS) PhD program for support. Both authors would also like to thank Vishaal Udandarao (University of T\"ubingen) for literature references helping in shaping this work.

{\small
\bibliographystyle{ieee_fullname}
\bibliography{egbib}
}

\newpage

\input{supp}



\end{document}

%% file: figures/first_page.tex
\begin{figure}
    \centering
    \includegraphics[width=1\linewidth]{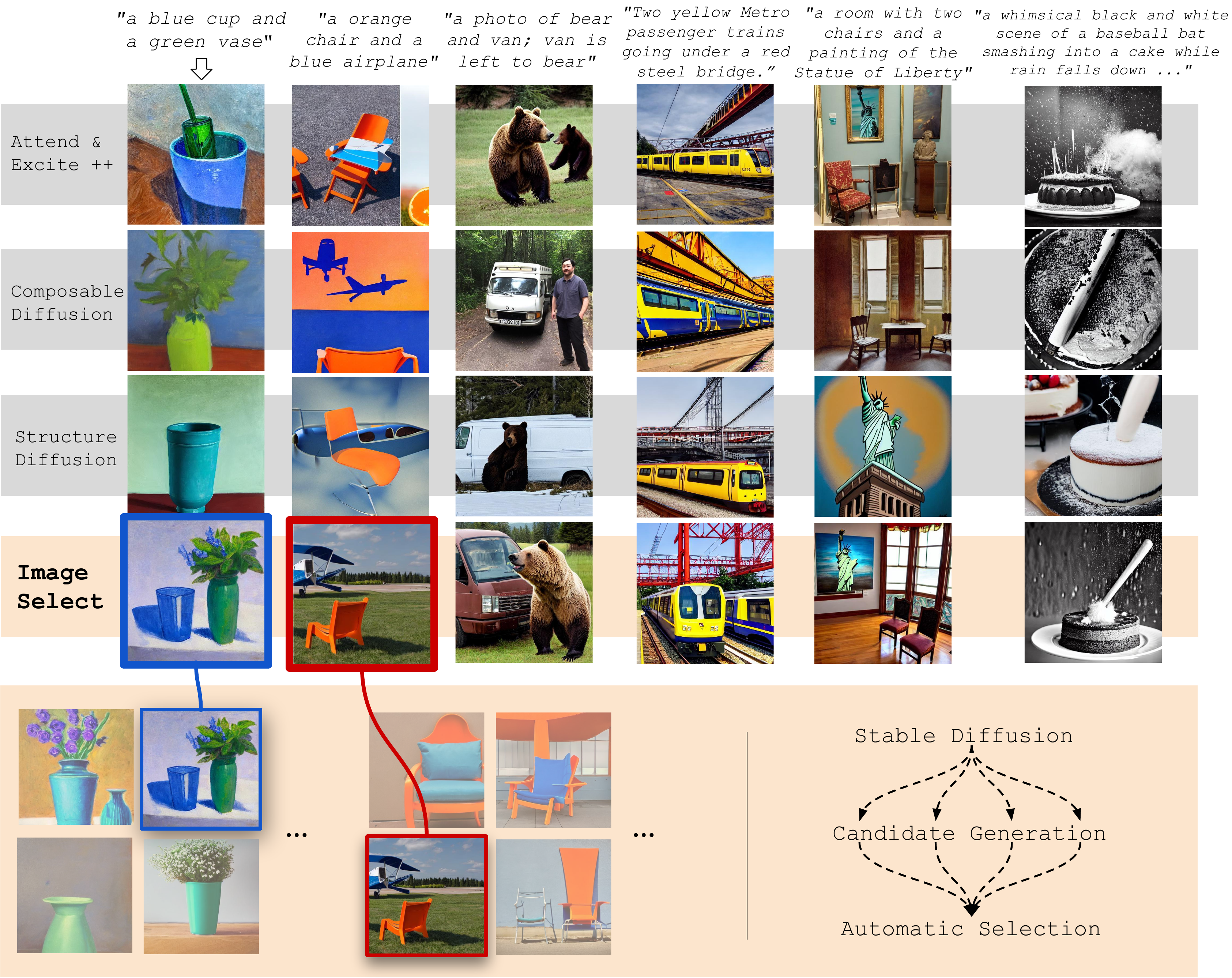}
    \caption{Our \texttt{ImageSelect} introduces automatic candidate selection to increase the faithfulness of a T2I generative model. We show that existing models are more faithful than assumed, and by simply querying them multiple times and selecting the most suitable image, we achieve significant improvements in T2I faithfulness, 
    without requiring to explicitly adapt the generative process.}
    \label{fig:first_page}
\end{figure}

%% file: figures/arch.tex
\begin{figure}
    \centering
    \includegraphics[width=\textwidth]{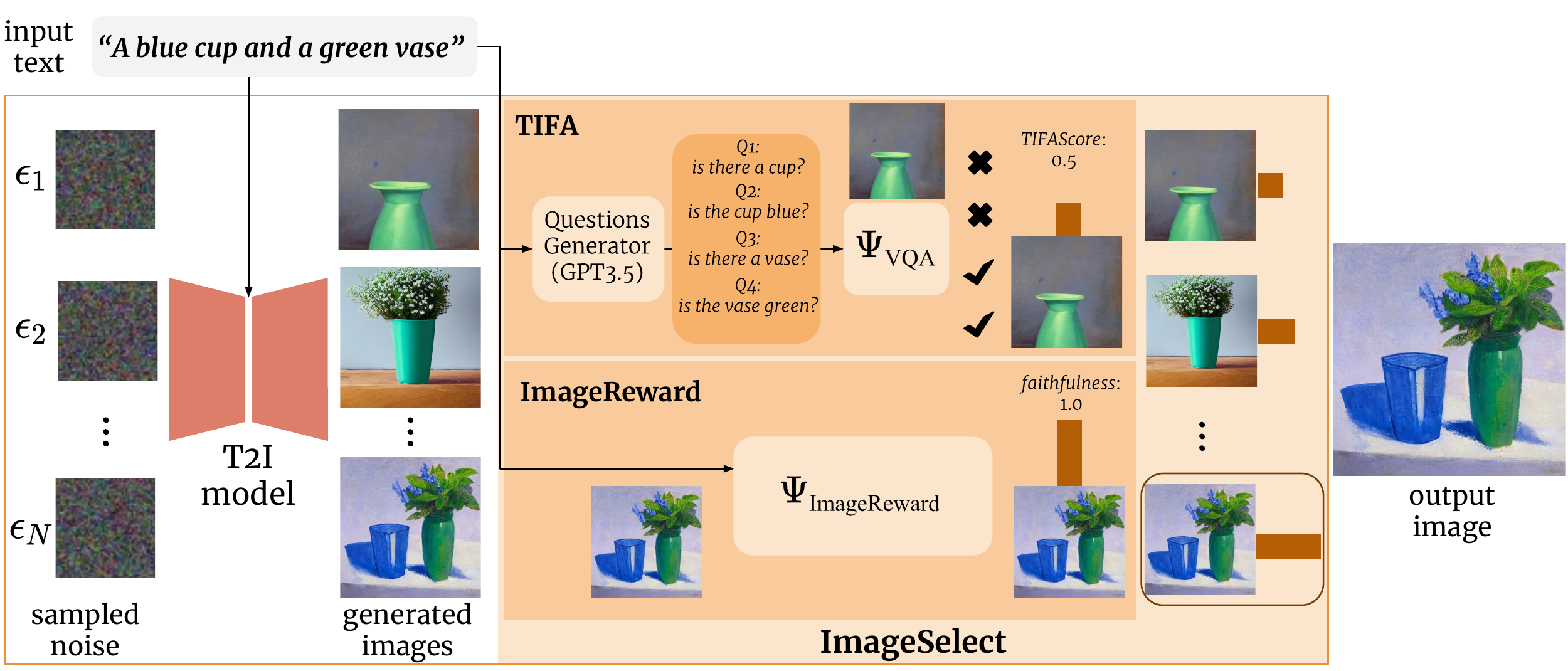}
    \caption{
    Given a text prompt and a set of latent starting points $\epsilon_i$, we generate corresponding candidate images with off-the-shelf T2I models s.a. Stable Diffusion. A scoring mechanism then assigns faithfulness scores per image, with the highest scoring one simply selected as the final output.
    }
    \label{fig:method}
\vspace{-5pt}
\end{figure}

%% file: tables/diverse_prompts.tex

\begin{wraptable}{r}{0.5\textwidth}
    \centering
    \vspace{-7mm}
    \caption{Summary statistics in our \texttt{diverse-1k} dataset. For furter details, see supplementary.}    
    \begin{tabular}{l|p{0.5\linewidth}|p{0.1\linewidth}}
    \toprule
         \textbf{Sources}$\downarrow$ & Subsets & Count \\ \midrule
         \multirow{4}{*}{HRS \cite{hrs}} & Bias, Spatial, Counting, Emotion, Size, Fairness, Length, Color, Synthetic & \multirow{3}{*}{38} \\
         & Writing & 36 \\ 
         \midrule
         \multirow{2}{*}{StrD \cite{feng2023structured}} & ABC & 127 \\
         & CC & 125  \\
         \midrule
         TIFA \cite{hu2023tifa} & N/A & 381 \\
         \midrule
         \multicolumn{3}{c}{\textbf{Total:} 1011}\\
    \bottomrule
    \end{tabular}
       \vspace{-2mm}
    \label{tab:diverse_prompts}
\end{wraptable}


%% file: figures/metric_comp.tex
\begin{figure}
    \centering
    \includegraphics[width=1\textwidth]{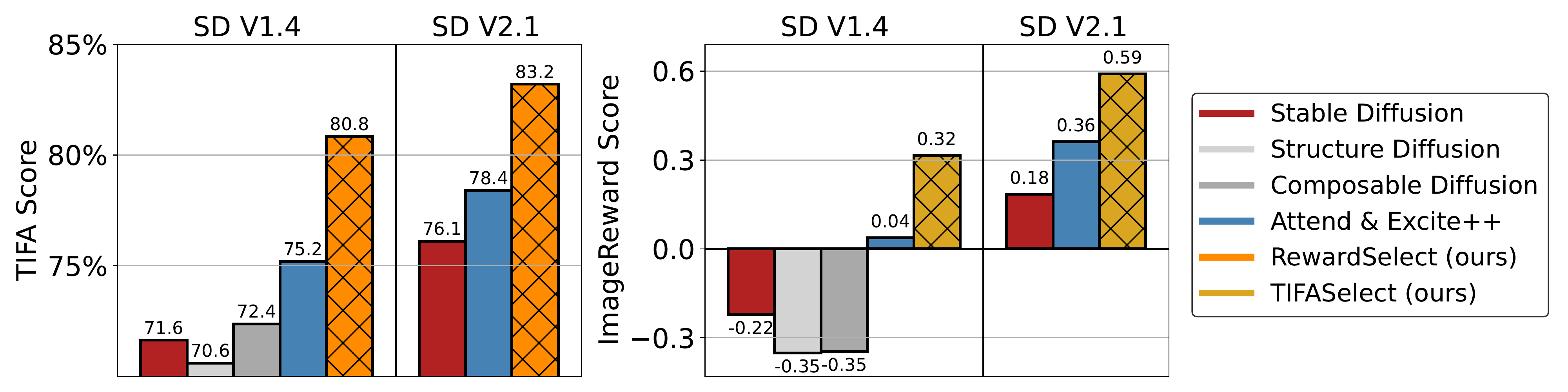}
    \vspace{-7mm}
    \caption{Quantitative results for baselines and \texttt{ImageSelect} on \texttt{diverse-1k}. For Stable Diffusion 1.4 and 2.1, \texttt{ImageSelect} outperforms all, irrespective of the selection and evaluation metric. }
    \label{fig:metrics_comp}
\end{figure}

%% file: figures/detail_comp.tex
\begin{wrapfigure}{r}{0.5\textwidth}
    \vspace{-20pt}
    \centering
    \includegraphics[width=0.5\textwidth]{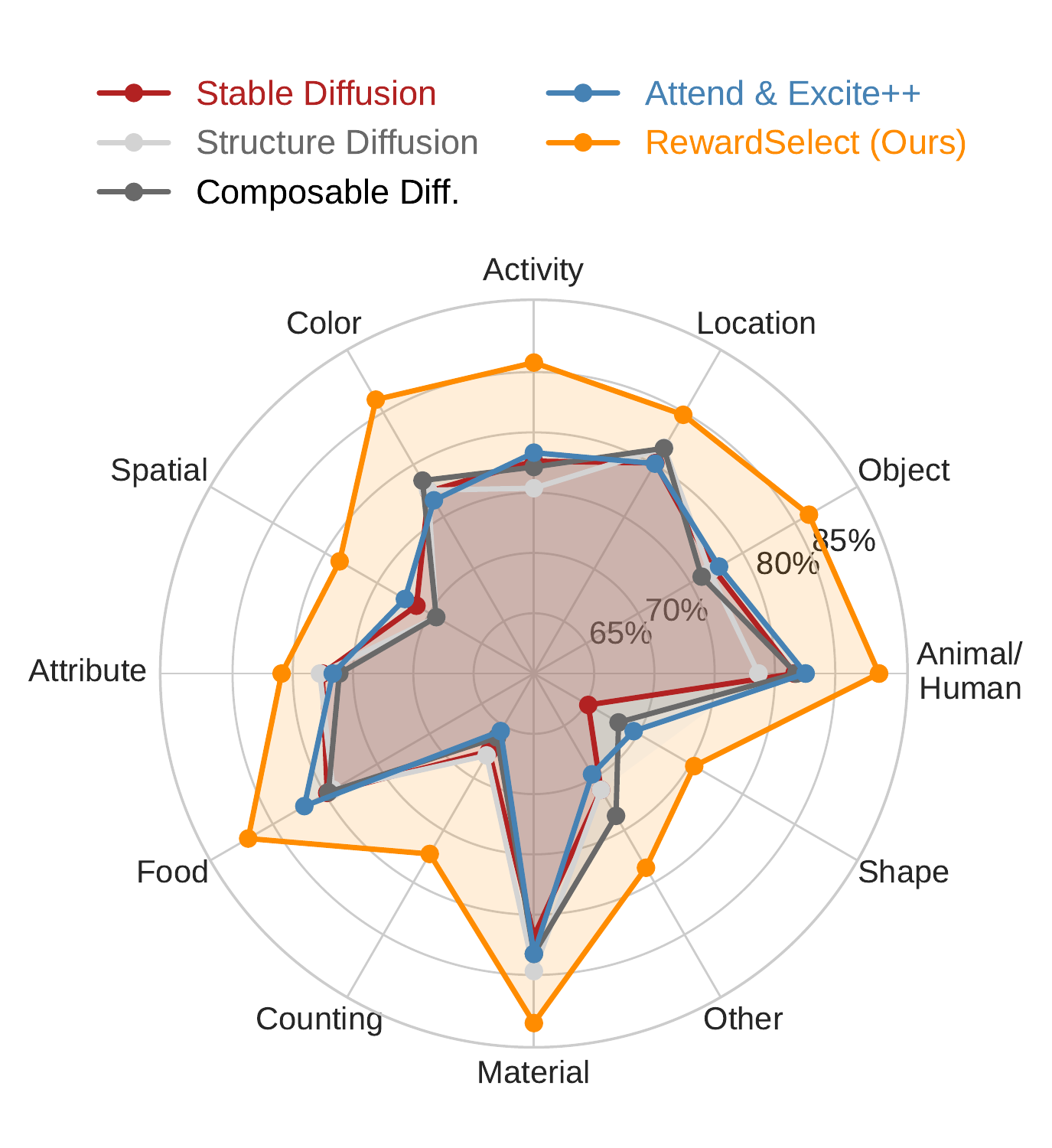}
    \vspace{-20pt}    
    \caption{\texttt{RewardSelect} offers improved faithfulness across faithfulness categories as used in \cite{hu2023tifa}}
    \label{fig:detail_comp}
    \vspace{-8pt}
\end{wrapfigure}

%% file: tables/ground_truth_comp.tex
\begin{wraptable}{r}{0.5\textwidth}
    \vspace{-5pt}
    \centering
    \vspace{-5mm}
    \caption{Faithfulness comparison with our \texttt{RewardSelect} (\texttt{RS}) using the TIFA-score on the ground-truth MS-COCO image-caption pairs. Our \texttt{RS} closes the gap  with GT=$89.09\%$ in faithfulness.} 
    \vspace{5pt}
    \begin{tabular}{l|lll}
    \toprule
         \multirow{2}{*}{V1.4}&  SD  & A\&E++ & \texttt{RS}  \\
          & $82.69\%$ & $82.04\%$ & $88.69\%$ \\ \hline
          \multirow{2}{*}{V2.1} & SD  & A\&E++  & \texttt{RS}  \\ 
          & $85.28\%$ & $85.87\%$ & $\mathbf{89.85}\%$\\ 
    \bottomrule
    \end{tabular}
       \vspace{-2mm}
    \label{tab:ground_truth_comp}
    \vspace{-5pt}
\end{wraptable}


%% file: figures/num_imgs.tex
\begin{wrapfigure}{r}{0.5\textwidth}
\vspace{-15pt}
    \centering
    \includegraphics[width=0.5\textwidth]{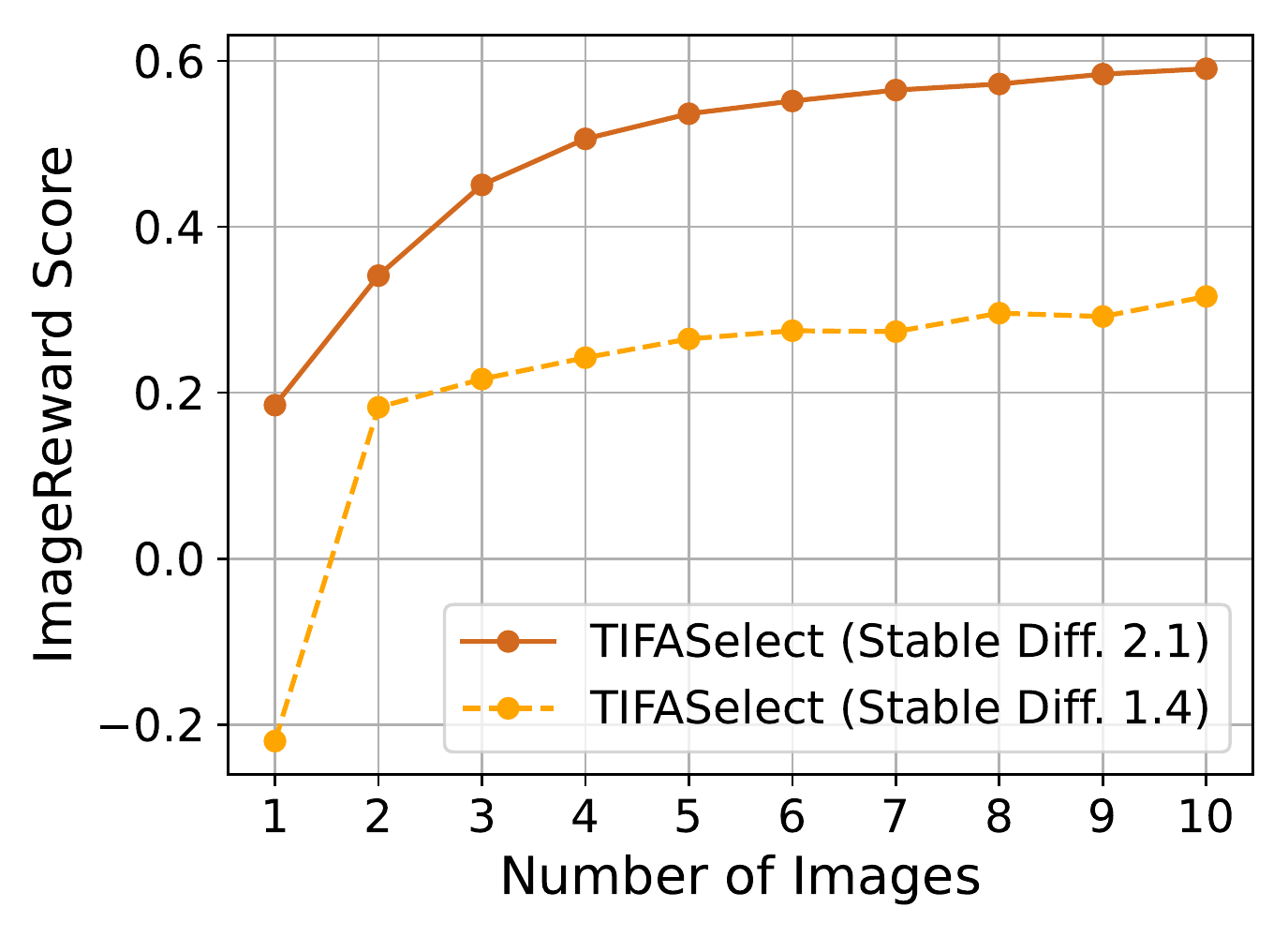}
    \vspace{-20pt}
    \caption{Faithfulness increases with number of candidate images per prompt to select from.}
    \label{fig:seed_comp}
\vspace{-10pt}
\end{wrapfigure}

%% file: figures/human_study.tex
\begin{figure}
    \centering
    \includegraphics[width=1\textwidth]{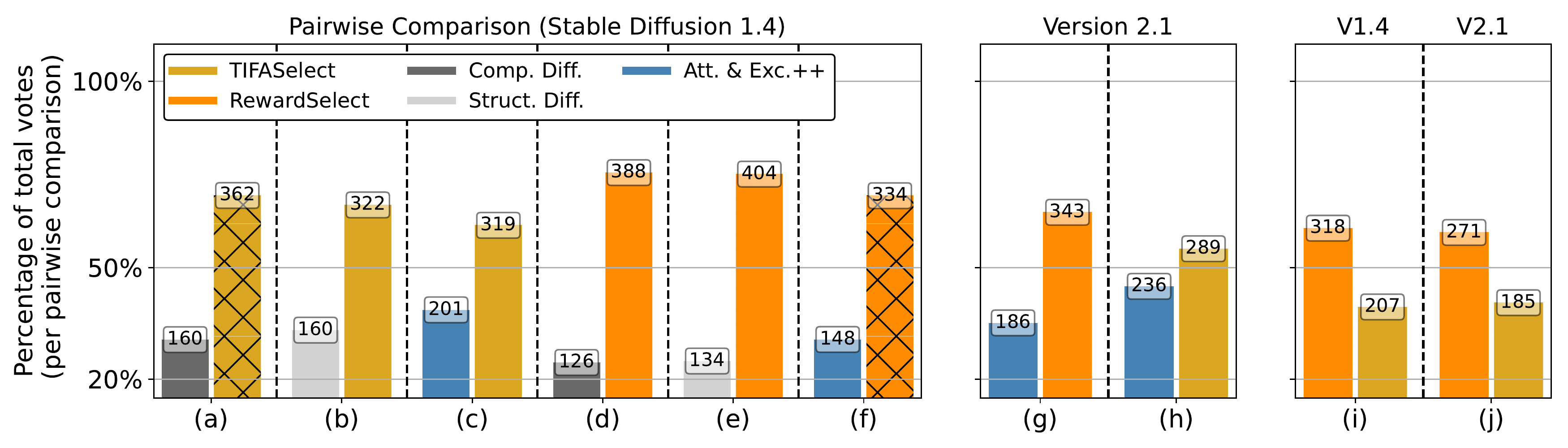}
    \vspace{-8mm}
    \caption{Performing human faithfulness comparisons between baselines and \texttt{ImageSelect} shows \texttt{ImageSelect} being preferred in the majority of cases for prompts from \texttt{diverse-1k}.}
    \vspace{-5mm}
    \label{fig:human_study}
\end{figure}

%% file: tables/rel_preferences.tex
\begin{wraptable}{r}{0.55\textwidth}
    \vspace{-4pt}
    \centering
    \caption{Relative improvements of \texttt{ImageSelect} approaches over faithfulness baselines. Human participants are in parts $\times2$  or even $\times3$ as likely to find \texttt{RewardSelect} images more faithful to the prompt. Even against our updated, automatic variation of A\&E, selection preference are in parts $>\times2$. Finally, comparing selection methods, we find the learned \texttt{RewardSelect} approaches to generally outperform \texttt{TIFASelect} which decomposes the matching tasks.}   
    \setlength{\tabcolsep}{3pt}
\renewcommand{\arraystretch}{1.1}
    \resizebox{\linewidth}{!}{
    \begin{tabular}{l|l|ccc|c}
    \toprule
        \multicolumn{2}{l|}{\textbf{Versus} $\rightarrow$} & CD \cite{liu2022compositional} & SD \cite{feng2023structured} & A\&E & \texttt{TIFASelect}\\
        \midrule
        \multirow{2}{*}{\textbf{V1.4}} & \texttt{TIFASelect} & $126.3$ & $101.24$ & $58.7$ & $\times$\\
        & \texttt{RewardSelect} & $207.9$ & $201.5$ & $125.7$ & $53.6$\\
        \midrule
        \multirow{2}{*}{\textbf{V2.1}} & \texttt{TIFASelect} & $\times$ & $\times$ & $22.5$ & $\times$\\
        & \texttt{RewardSelect} & $\times$ & $\times$ & $84.4$ & $46.5$\\        
    \bottomrule
    \end{tabular}}
    \label{tab:user_study}
\vspace{-7pt}
\end{wraptable}

%% file: figures/more_examples.tex
\begin{figure}
    \centering
    \includegraphics[width=1\linewidth]{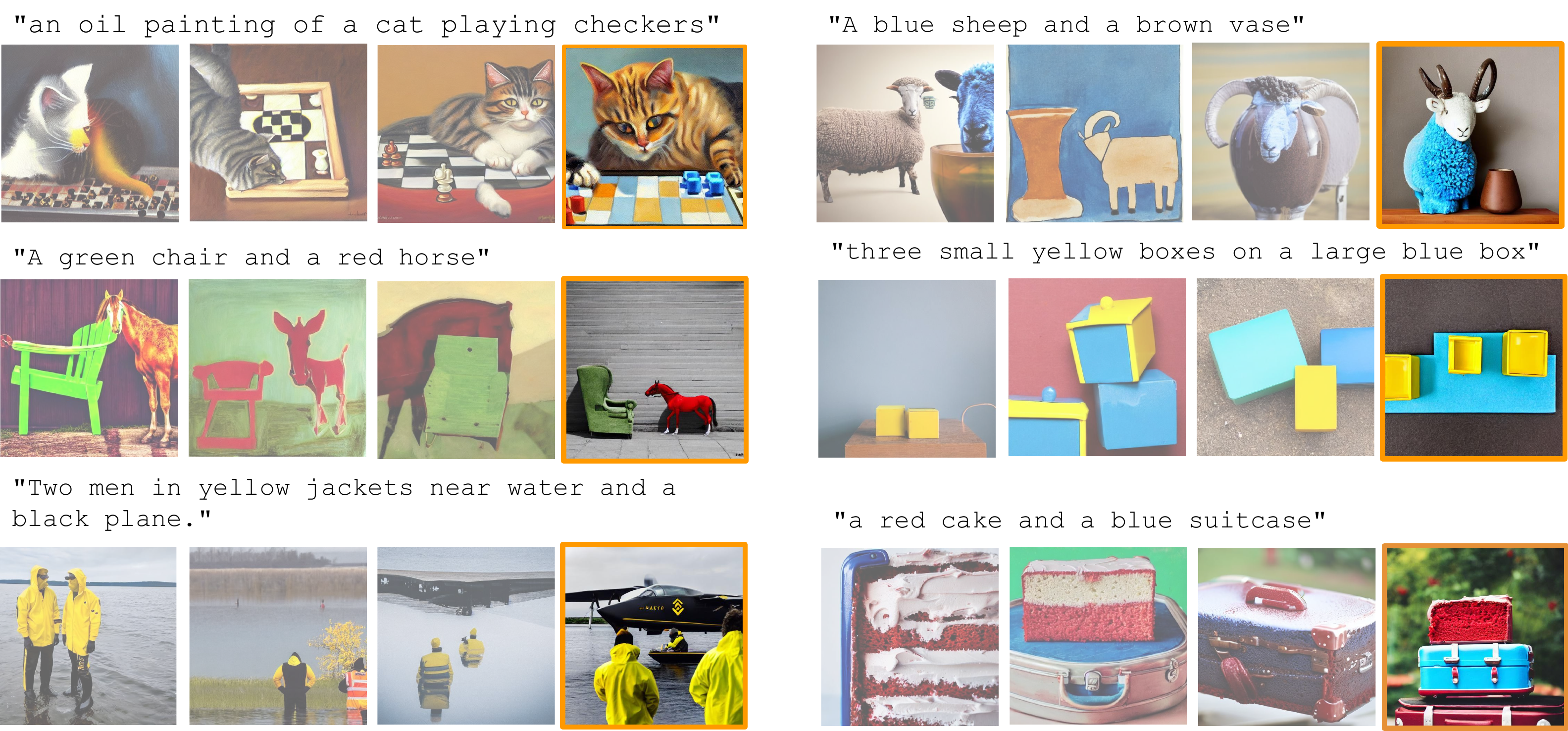}
    \caption{\textit{Additional Examples} highlighting favorable faithfulness of \texttt{ImageSelect} (rightmost) compared to Attend-and-Excite++, Composable Diffusion \cite{liu2022compositional} and Structure Diffusion \cite{feng2023structured}.}
    \label{fig:more_examples}
\vspace{-5pt}
\end{figure}

%% file: figures/failure_cases.tex
\begin{figure}
    \centering
    \includegraphics[width=1\linewidth]{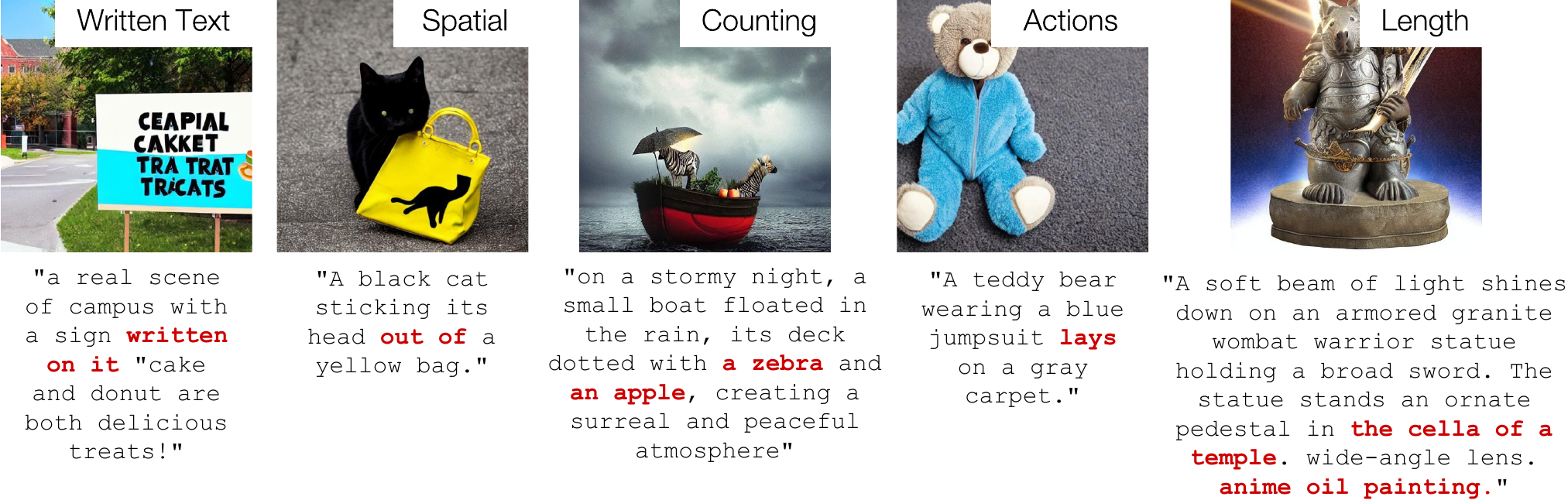}
    \vspace{-5mm}
    \caption{\textit{Qualitative Failure Cases.} Despite significantly improving faithfulness, \texttt{ImageSelect} can not fully account for fundamental shortcomings. 
    Details on faithfulness categories, see e.g. Fig.~\ref{fig:detail_comp}.}
    \vspace{-2mm}
    \label{fig:failure_cases}
\end{figure}

%% file: supp.tex
\par\noindent\rule{\textwidth}{1.4pt}

\Large{\textbf{Supplementary}}

\Large{\textbf{If at First You Don't Succeed, Try, Try Again:\\ Faithful Diffusion-based Text-to-Image Generation by Selection}}

\par\noindent\rule{\textwidth}{1.4pt}

\vspace{10pt}

\normalsize



\section{Implementation Details}
We conduct all experiments using the PyTorch framework \cite{pytorch} on a high-performance compute cluster comprising NVIDIA 2080Ti GPUs. We use the publicly available implementations of \href{https://github.com/yuval-alaluf/Attend-and-Excite}{Attend-and-Excite}  \href{https://github.com/weixi-feng/Structured-Diffusion-Guidance}{Structure Diffusion}, and \href{https://github.com/energy-based-model/Compositional-Visual-Generation-with-Composable-Diffusion-Models-PyTorch}{Composable Diffusion}. We were unable to benchmark Space-Time-Attention due to its high computational requirements.

\section{User Study}
To compare our \texttt{ImageSelect} variants on faithfulness leveraging human feedback, we set up a simple study, in which participants are given a prompt, taken from the \texttt{diverse-1k} dataset, and two associated images. We set this up as shown in Fig.~\ref{fig:study_gui}, where one image is taken from one of the baseline methods, and one from a respective \texttt{ImageSelect} variant. The position of each in the GUI is determined at random for each selection task.
For Stable Diffusion V1.4, these baselines are Structure Diffusion \cite{feng2023structured}, Composable Diffusion \cite{liu2022compositional} or Attend-and-Excite++. For Stable Diffusion V2.1, we utilize Attend-and-Excite++. In addition to that, we also compare \texttt{ImageSelect} variants, \texttt{TIFASelect} and \texttt{RewardSelect}, across both generations of Stable Diffusion.
Before participation, each user is tasked to select the image they think most faithfully reflects the textural prompt.

The complete study is conducted through a web link, which is publicly shared and distributed. Each user participates entirely voluntarily and can start and end their participation whenever desired. Overall, we collect data for one week, aggregating 5093 selections for all pairwise comparisons, and distributed over 68 users.

\section{Additional Qualitative Results}
In this section, we provide additional qualitative comparisons in Figure \ref{fig:more_examples_supp} for Stable Diffusion V1.4, and Figure \ref{fig:more_examples_supp_v2} for Stable Diffusion V2.1, where we compare against Attend-and-Excite++, to extend those shown in the main paper. Reflective of both quantitative results and human study evaluations, we find clear qualitative evidence of increased faithfulness when leveraging automatic candidate selection through \texttt{ImageSelect} variants (in these cases \texttt{RewardSelect}).

\input{figures/study_gui}

\input{figures/more_examples_supp}

%% file: figures/study_gui.tex
\begin{figure}
    \centering
    \includegraphics[width=0.7\textwidth]{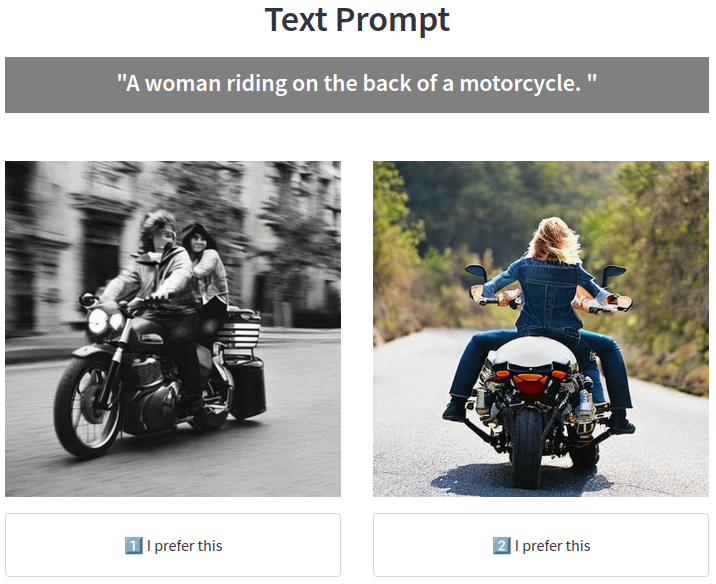}
    \caption{
        User interface for our human faithfulness study. A user is presented with the simple task of selecting which presented image more faithfully represents the given text prompt. We opted for binary comparisons as these tasks are easiest to evaluate for human users. Images presented are randomly selected from method pairs, with one baseline method (i.e. Compositional Diffusion \cite{liu2022compositional}, Structured Diffusion \cite{feng2023structured} or Attend-and-Excite~\cite{chefer2023attendandexcite}) and an \texttt{ImageSelect} variant. Results are collected anonymously.
    }
    \label{fig:study_gui}
\end{figure}

%% file: figures/more_examples_supp.tex
\begin{figure}[t]
    \centering
    \includegraphics[width=1\linewidth]{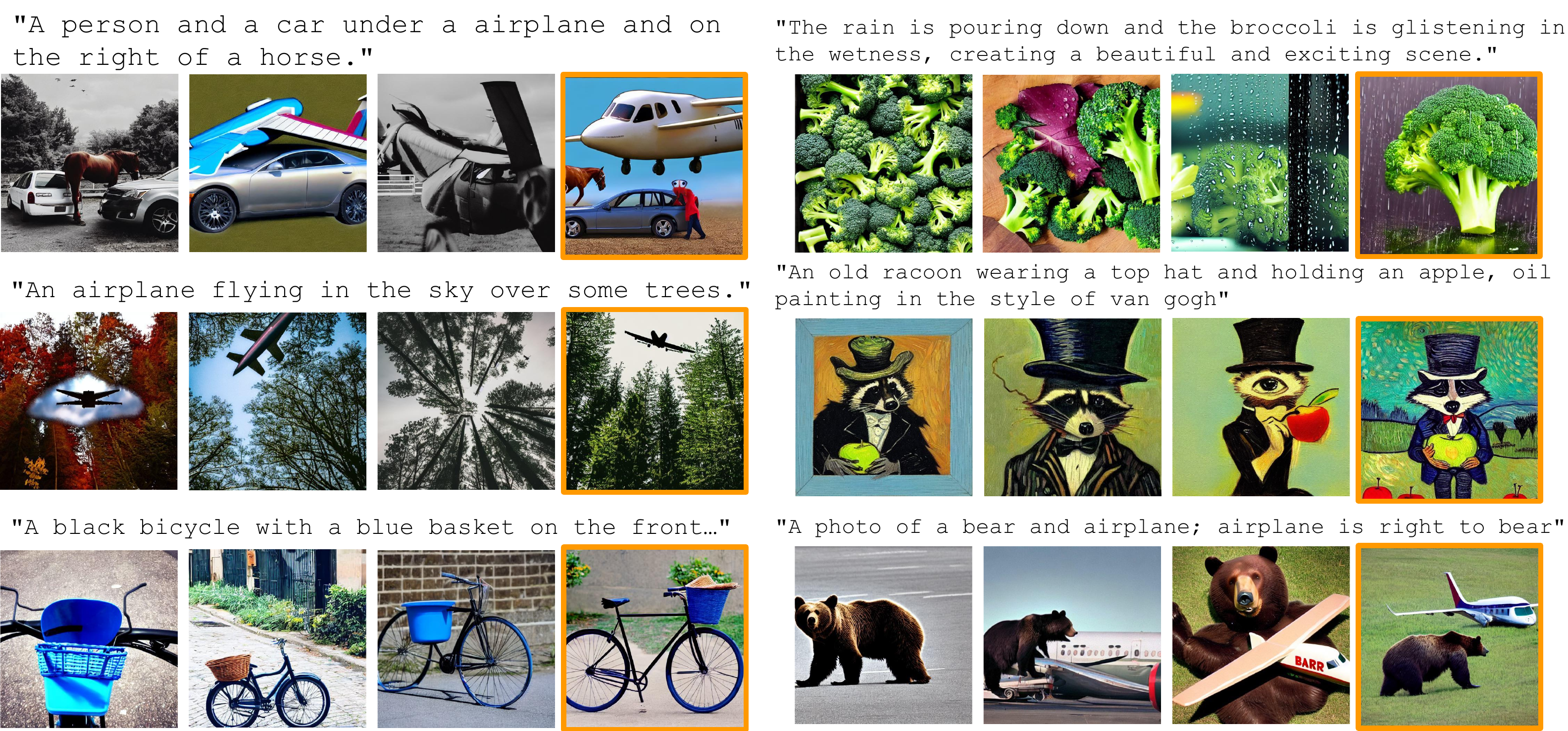}
    \caption{\textit{Additional Examples} highlighting favorable faithfulness of \texttt{ImageSelect} (rightmost) compared to Attend-and-Excite++, Composable Diffusion \cite{liu2022compositional} and Structure Diffusion \cite{feng2023structured}.}
    \label{fig:more_examples_supp}
\end{figure}
\begin{figure}[t]
    \centering
    \includegraphics[width=1\linewidth]{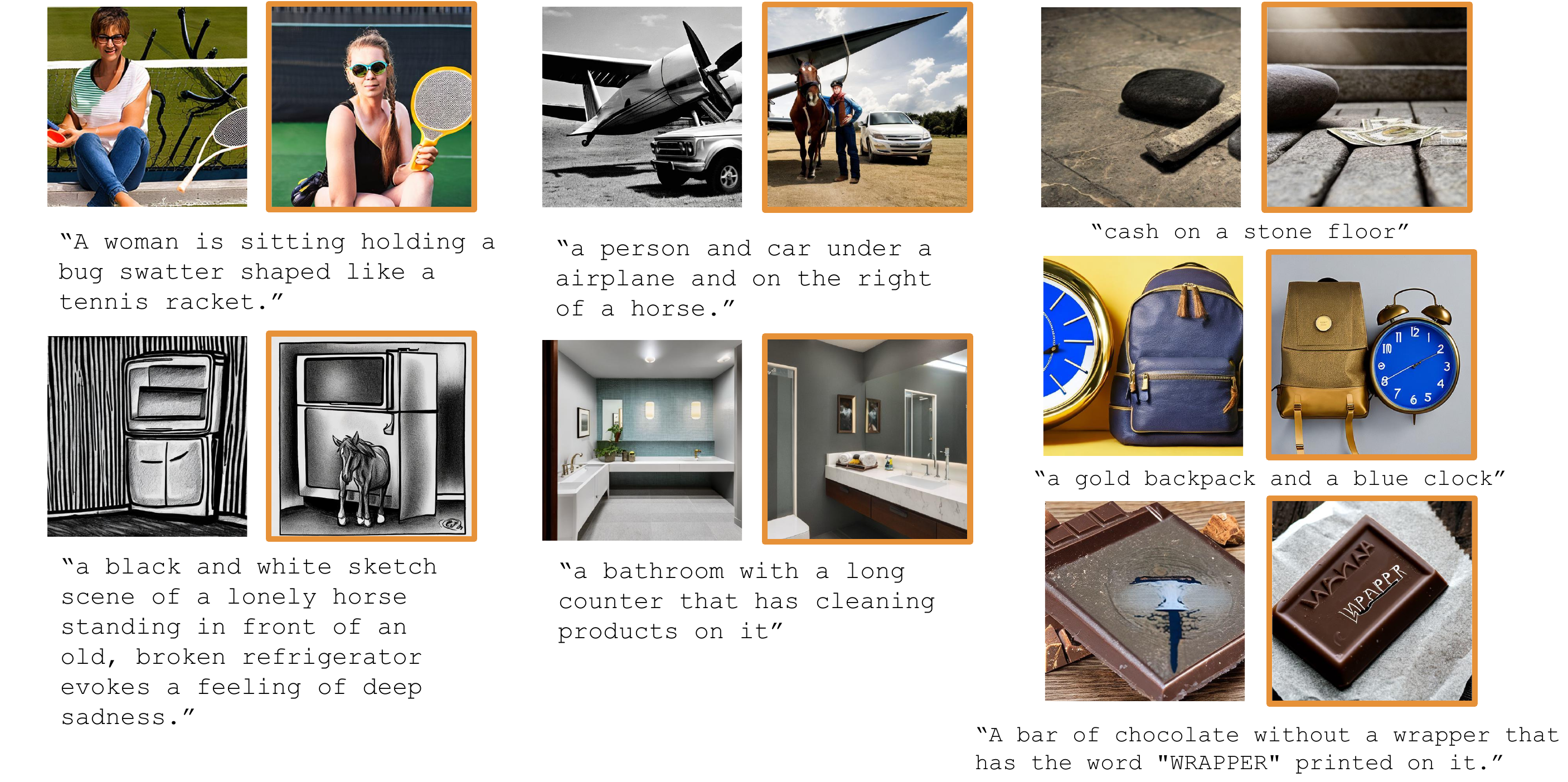}
    \caption{\textit{Additional Examples} for Stable Diffusion V2.1, comparing Attend-and-Excite++ and \texttt{RewardSelect} (right). While the change in model generation offers additional improvements in faithfulness, we find the additional use of \texttt{RewardSelect} to still offer notable benefits, which is also clearly reflected qualitatively.}
    \label{fig:more_examples_supp_v2}
\end{figure}

%% file: arxivmain.bbl
\begin{thebibliography}{10}\itemsep=-1pt

\bibitem{alaniz2022semantic}
Stephan Alaniz, Thomas Hummel, and Zeynep Akata.
\newblock Semantic image synthesis with semantically coupled vq-model.
\newblock {\em arXiv preprint arXiv:2209.02536}, 2022.

\bibitem{anderson2016spice}
Peter Anderson, Basura Fernando, Mark Johnson, and Stephen Gould.
\newblock Spice: Semantic propositional image caption evaluation.
\newblock In {\em ECCV}, 2016.

\bibitem{vqa}
Stanislaw Antol, Aishwarya Agrawal, Jiasen Lu, Margaret Mitchell, Dhruv Batra,
  C~Lawrence Zitnick, and Devi Parikh.
\newblock Vqa: Visual question answering.
\newblock In {\em ICCV}, 2015.

\bibitem{hrs}
Eslam~Mohamed Bakr, Pengzhan Sun, Xiaoqian Shen, Faizan~Farooq Khan, Li~Erran
  Li, and Mohamed Elhoseiny.
\newblock Hrs-bench: Holistic, reliable and scalable benchmark for
  text-to-image models.
\newblock {\em arXiv preprint arXiv:2304.05390}, 2023.

\bibitem{banerjee2005meteor}
Satanjeev Banerjee and Alon Lavie.
\newblock {METEOR}: An automatic metric for {MT} evaluation with improved
  correlation with human judgments.
\newblock In {\em ACL-W}, 2005.

\bibitem{bias2}
Hritik Bansal, Da Yin, Masoud Monajatipoor, and Kai-Wei Chang.
\newblock How well can text-to-image generative models understand ethical
  natural language interventions?
\newblock In {\em EMNLP}, 2022.

\bibitem{bommasani2021opportunities}
Rishi Bommasani, Drew~A Hudson, Ehsan Adeli, Russ Altman, Simran Arora, Sydney
  von Arx, Michael~S Bernstein, Jeannette Bohg, Antoine Bosselut, Emma
  Brunskill, et~al.
\newblock On the opportunities and risks of foundation models.
\newblock {\em arXiv preprint arXiv:2108.07258}, 2021.

\bibitem{gpt3}
Tom Brown, Benjamin Mann, Nick Ryder, Melanie Subbiah, Jared~D Kaplan, Prafulla
  Dhariwal, Arvind Neelakantan, Pranav Shyam, Girish Sastry, Amanda Askell,
  et~al.
\newblock Language models are few-shot learners.
\newblock {\em NeurIPS}, 2020.

\bibitem{changpinyo2022all}
Soravit Changpinyo, Doron Kukliansky, Idan Szpektor, Xi Chen, Nan Ding, and
  Radu Soricut.
\newblock All you may need for vqa are image captions.
\newblock In {\em NAACL}, 2022.

\bibitem{chefer2023attendandexcite}
Hila Chefer, Yuval Alaluf, Yael Vinker, Lior Wolf, and Daniel Cohen-Or.
\newblock Attend-and-excite: Attention-based semantic guidance for
  text-to-image diffusion models.
\newblock In {\em SIGGRAPH}, 2023.

\bibitem{chefer2021generic}
Hila Chefer, Shir Gur, and Lior Wolf.
\newblock Generic attention-model explainability for interpreting bi-modal and
  encoder-decoder transformers.
\newblock In {\em ICCV}, 2021.

\bibitem{cococaptions}
Xinlei Chen, Hao Fang, Tsung-Yi Lin, Ramakrishna Vedantam, Saurabh Gupta, Piotr
  Doll{\'a}r, and C~Lawrence Zitnick.
\newblock Microsoft coco captions: Data collection and evaluation server.
\newblock {\em arXiv preprint arXiv:1504.00325}, 2015.

\bibitem{dalleeval}
Jaemin Cho, Abhay Zala, and Mohit Bansal.
\newblock Dall-eval: Probing the reasoning skills and social biases of
  text-to-image generative transformers.
\newblock {\em arXiv preprint arXiv:2202.04053}, 2022.

\bibitem{openairlhf}
Paul~F Christiano, Jan Leike, Tom Brown, Miljan Martic, Shane Legg, and Dario
  Amodei.
\newblock Deep reinforcement learning from human preferences.
\newblock {\em NIPS}, 2017.

\bibitem{vqgan}
Katherine Crowson, Stella Biderman, Daniel Kornis, Dashiell Stander, Eric
  Hallahan, Louis Castricato, and Edward Raff.
\newblock Vqgan-clip: Open domain image generation and editing with natural
  language guidance.
\newblock In {\em ECCV}, 2022.

\bibitem{dhariwal2021diffusion}
Prafulla Dhariwal and Alexander Nichol.
\newblock Diffusion models beat gans on image synthesis.
\newblock {\em NeurIPS}, 2021.

\bibitem{elBanani2023lang}
Mohamed El~Banani, Karan Desai, and Justin Johnson.
\newblock Learning visual representations via language-guided sampling.
\newblock In {\em Proceedings of the IEEE/CVF Conference on Computer Vision and
  Pattern Recognition (CVPR)}, pages 19208--19220, June 2023.

\bibitem{esser2021taming}
Patrick Esser, Robin Rombach, and Bjorn Ommer.
\newblock Taming transformers for high-resolution image synthesis.
\newblock In {\em CVPR}, 2021.

\bibitem{feng2023structured}
Weixi Feng, Xuehai He, Tsu-Jui Fu, Varun Jampani, Arjun~Reddy Akula, Pradyumna
  Narayana, Sugato Basu, Xin~Eric Wang, and William~Yang Wang.
\newblock Training-free structured diffusion guidance for compositional
  text-to-image synthesis.
\newblock In {\em ICLR}, 2023.

\bibitem{makeascene}
Oran Gafni, Adam Polyak, Oron Ashual, Shelly Sheynin, Devi Parikh, and Yaniv
  Taigman.
\newblock Make-a-scene: Scene-based text-to-image generation with human priors.
\newblock In {\em ECCV}, 2022.

\bibitem{gokhale2022benchmarking}
Tejas Gokhale, Hamid Palangi, Besmira Nushi, Vibhav Vineet, Eric Horvitz, Ece
  Kamar, Chitta Baral, and Yezhou Yang.
\newblock Benchmarking spatial relationships in text-to-image generation.
\newblock {\em arXiv preprint arXiv:2212.10015}, 2022.

\bibitem{goodfellow2014gan}
Ian Goodfellow, Jean Pouget-Abadie, Mehdi Mirza, Bing Xu, David Warde-Farley,
  Sherjil Ozair, Aaron Courville, and Yoshua Bengio.
\newblock Generative adversarial nets.
\newblock In {\em NIPS}, 2014.

\bibitem{ogrlfh}
Shane Griffith, Kaushik Subramanian, Jonathan Scholz, Charles~L Isbell, and
  Andrea~L Thomaz.
\newblock Policy shaping: Integrating human feedback with reinforcement
  learning.
\newblock {\em NIPS}, 2013.

\bibitem{gu2022vector}
Shuyang Gu, Dong Chen, Jianmin Bao, Fang Wen, Bo Zhang, Dongdong Chen, Lu Yuan,
  and Baining Guo.
\newblock Vector quantized diffusion model for text-to-image synthesis.
\newblock In {\em CVPR}, 2022.

\bibitem{hertz2022prompt}
Amir Hertz, Ron Mokady, Jay Tenenbaum, Kfir Aberman, Yael Pritch, and Daniel
  Cohen-Or.
\newblock Prompt-to-prompt image editing with cross attention control.
\newblock {\em arXiv preprint arXiv:2208.01626}, 2022.

\bibitem{hessel2021clipscore}
Jack Hessel, Ari Holtzman, Maxwell Forbes, Ronan~Le Bras, and Yejin Choi.
\newblock Clipscore: A reference-free evaluation metric for image captioning.
\newblock In {\em EMNLP}, 2021.

\bibitem{ho2020denoising}
Jonathan Ho, Ajay Jain, and Pieter Abbeel.
\newblock Denoising diffusion probabilistic models.
\newblock In {\em NeurIPS}, 2020.

\bibitem{hu2023tifa}
Yushi Hu, Benlin Liu, Jungo Kasai, Yizhong Wang, Mari Ostendorf, Ranjay
  Krishna, and Noah~A Smith.
\newblock Tifa: Accurate and interpretable text-to-image faithfulness
  evaluation with question answering.
\newblock {\em arXiv preprint arXiv:2303.11897}, 2023.

\bibitem{ilharco2021openclip}
Gabriel Ilharco, Mitchell Wortsman, Ross Wightman, Cade Gordon, Nicholas
  Carlini, Rohan Taori, Achal Dave, Vaishaal Shankar, Hongseok Namkoong, John
  Miller, Hannaneh Hajishirzi, Ali Farhadi, and Ludwig Schmidt.
\newblock Openclip, July 2021.
\newblock If you use this software, please cite it as below.

\bibitem{kang2023scaling}
Minguk Kang, Jun-Yan Zhu, Richard Zhang, Jaesik Park, Eli Shechtman, Sylvain
  Paris, and Taesung Park.
\newblock Scaling up gans for text-to-image synthesis.
\newblock {\em arXiv preprint arXiv:2303.05511}, 2023.

\bibitem{kingma2014vae}
Diederik~P. Kingma and Max Welling.
\newblock {Auto-Encoding Variational Bayes}.
\newblock In {\em ICLR}, 2014.

\bibitem{lee2022autoregressive}
Doyup Lee, Chiheon Kim, Saehoon Kim, Minsu Cho, and Wook-Shin Han.
\newblock Autoregressive image generation using residual quantization.
\newblock In {\em Proceedings of the IEEE/CVF Conference on Computer Vision and
  Pattern Recognition}, pages 11523--11532, 2022.

\bibitem{lee2023aligning}
Kimin Lee, Hao Liu, Moonkyung Ryu, Olivia Watkins, Yuqing Du, Craig Boutilier,
  Pieter Abbeel, Mohammad Ghavamzadeh, and Shixiang~Shane Gu.
\newblock Aligning text-to-image models using human feedback.
\newblock {\em arXiv preprint arXiv:2302.12192}, 2023.

\bibitem{mplug}
Chenliang Li, Haiyang Xu, Junfeng Tian, Wei Wang, Ming Yan, Bin Bi, Jiabo Ye,
  Hehong Chen, Guohai Xu, Zheng Cao, et~al.
\newblock mplug: Effective and efficient vision-language learning by
  cross-modal skip-connections.
\newblock In {\em EMNLP}, 2022.

\bibitem{blip2}
Junnan Li, Dongxu Li, Silvio Savarese, and Steven Hoi.
\newblock Blip-2: Bootstrapping language-image pre-training with frozen image
  encoders and large language models.
\newblock In {\em ICML}, 2023.

\bibitem{blip}
Junnan Li, Dongxu Li, Caiming Xiong, and Steven Hoi.
\newblock Blip: Bootstrapping language-image pre-training for unified
  vision-language understanding and generation.
\newblock In {\em International Conference on Machine Learning}, 2022.

\bibitem{mscoco}
Tsung-Yi Lin, Michael Maire, Serge Belongie, James Hays, Pietro Perona, Deva
  Ramanan, Piotr Doll{\'a}r, and C~Lawrence Zitnick.
\newblock Microsoft coco: Common objects in context.
\newblock In {\em ECCV}, 2014.

\bibitem{vsr}
Fangyu Liu, Guy Emerson, and Nigel Collier.
\newblock Visual spatial reasoning.
\newblock {\em TACL}, 2023.

\bibitem{liu2022compositional}
Nan Liu, Shuang Li, Yilun Du, Antonio Torralba, and Joshua~B. Tenenbaum.
\newblock Compositional visual generation with composable diffusion models.
\newblock In {\em ECCV}, 2022.

\bibitem{spelling}
Rosanne Liu, Dan Garrette, Chitwan Saharia, William Chan, Adam Roberts, Sharan
  Narang, Irina Blok, RJ Mical, Mohammad Norouzi, and Noah Constant.
\newblock Character-aware models improve visual text rendering.
\newblock {\em arXiv preprint arXiv:2212.10562}, 2022.

\bibitem{loper2002nltk}
Edward Loper and Steven Bird.
\newblock Nltk: The natural language toolkit.
\newblock {\em arXiv preprint cs/0205028}, 2002.

\bibitem{mansimov2015generating}
Elman Mansimov, Emilio Parisotto, Jimmy~Lei Ba, and Ruslan Salakhutdinov.
\newblock Generating images from captions with attention.
\newblock In {\em ICLR}, 2016.

\bibitem{bias1}
Ranjita Naik and Besmira Nushi.
\newblock Social biases through the text-to-image generation lens.
\newblock {\em arXiv preprint arXiv:2304.06034}, 2023.

\bibitem{glide}
Alex Nichol, Prafulla Dhariwal, Aditya Ramesh, Pranav Shyam, Pamela Mishkin,
  Bob McGrew, Ilya Sutskever, and Mark Chen.
\newblock Glide: Towards photorealistic image generation and editing with
  text-guided diffusion models.
\newblock {\em arXiv preprint arXiv:2112.10741}, 2021.

\bibitem{humaneval}
Mayu Otani, Riku Togashi, Yu Sawai, Ryosuke Ishigami, Yuta Nakashima, Esa
  Rahtu, Janne Heikkil{\"a}, and Shin'ichi Satoh.
\newblock Toward verifiable and reproducible human evaluation for text-to-image
  generation.
\newblock In {\em CVPR}, 2023.

\bibitem{paiss2023teaching}
Roni Paiss, Ariel Ephrat, Omer Tov, Shiran Zada, Inbar Mosseri, Michal Irani,
  and Tali Dekel.
\newblock Teaching clip to count to ten.
\newblock {\em arXiv preprint arXiv:2302.12066}, 2023.

\bibitem{bleu}
Kishore Papineni, Salim Roukos, Todd Ward, and Wei-Jing Zhu.
\newblock Bleu: a method for automatic evaluation of machine translation.
\newblock In {\em Proceedings of the 40th annual meeting of the Association for
  Computational Linguistics}, pages 311--318, 2002.

\bibitem{compt2i}
Dong~Huk Park, Samaneh Azadi, Xihui Liu, Trevor Darrell, and Anna Rohrbach.
\newblock Benchmark for compositional text-to-image synthesis.
\newblock In {\em NeurIPS Datasets and Benchmarks Track}, 2021.

\bibitem{pytorch}
Adam Paszke, Sam Gross, Francisco Massa, Adam Lerer, James Bradbury, Gregory
  Chanan, Trevor Killeen, Zeming Lin, Natalia Gimelshein, Luca Antiga, et~al.
\newblock Pytorch: An imperative style, high-performance deep learning library.
\newblock {\em Advances in neural information processing systems}, 32, 2019.

\bibitem{picard2021torch}
David Picard.
\newblock Torch. manual\_seed (3407) is all you need: On the influence of
  random seeds in deep learning architectures for computer vision.
\newblock {\em arXiv preprint arXiv:2109.08203}, 2021.

\bibitem{clip}
Alec Radford, Jong~Wook Kim, Chris Hallacy, Aditya Ramesh, Gabriel Goh,
  Sandhini Agarwal, Girish Sastry, Amanda Askell, Pamela Mishkin, Jack Clark,
  et~al.
\newblock Learning transferable visual models from natural language
  supervision.
\newblock In {\em ICML}, 2021.

\bibitem{t5}
Colin Raffel, Noam Shazeer, Adam Roberts, Katherine Lee, Sharan Narang, Michael
  Matena, Yanqi Zhou, Wei Li, and Peter~J Liu.
\newblock Exploring the limits of transfer learning with a unified text-to-text
  transformer.
\newblock {\em JMLR}, 2020.

\bibitem{dall-e2}
Aditya Ramesh, Prafulla Dhariwal, Alex Nichol, Casey Chu, and Mark Chen.
\newblock Hierarchical text-conditional image generation with clip latents.
\newblock {\em arXiv preprint arXiv:2204.06125}, 2022.

\bibitem{dall-e}
Aditya Ramesh, Mikhail Pavlov, Gabriel Goh, Scott Gray, Chelsea Voss, Alec
  Radford, Mark Chen, and Ilya Sutskever.
\newblock Zero-shot text-to-image generation.
\newblock In {\em ICML}. PMLR, 2021.

\bibitem{reed2016generative}
Scott Reed, Zeynep Akata, Xinchen Yan, Lajanugen Logeswaran, Bernt Schiele, and
  Honglak Lee.
\newblock Generative adversarial text to image synthesis.
\newblock In {\em ICML}, 2016.

\bibitem{reed2016learning}
Scott~E Reed, Zeynep Akata, Santosh Mohan, Samuel Tenka, Bernt Schiele, and
  Honglak Lee.
\newblock Learning what and where to draw.
\newblock {\em NIPS}, 2016.

\bibitem{rombach22ldm}
Robin Rombach, Andreas Blattmann, Dominik Lorenz, Patrick Esser, and Bj\"orn
  Ommer.
\newblock High-resolution image synthesis with latent diffusion models.
\newblock In {\em Proceedings of the IEEE/CVF Conference on Computer Vision and
  Pattern Recognition (CVPR)}, pages 10684--10695, June 2022.

\bibitem{unet}
Olaf Ronneberger, Philipp Fischer, and Thomas Brox.
\newblock U-net: Convolutional networks for biomedical image segmentation.
\newblock In {\em MICCAI}, 2015.

\bibitem{roth2022lang}
Karsten Roth, Oriol Vinyals, and Zeynep Akata.
\newblock Integrating language guidance into vision-based deep metric learning.
\newblock In {\em Proceedings of the IEEE/CVF Conference on Computer Vision and
  Pattern Recognition (CVPR)}, pages 16177--16189, June 2022.

\bibitem{imagen}
Chitwan Saharia, William Chan, Saurabh Saxena, Lala Li, Jay Whang, Emily~L
  Denton, Kamyar Ghasemipour, Raphael Gontijo~Lopes, Burcu Karagol~Ayan, Tim
  Salimans, et~al.
\newblock Photorealistic text-to-image diffusion models with deep language
  understanding.
\newblock {\em NeurIPS}, 35, 2022.

\bibitem{sauer2023stylegant}
Axel Sauer, Tero Karras, Samuli Laine, Andreas Geiger, and Timo Aila.
\newblock Stylegan-t: Unlocking the power of gans for fast large-scale
  text-to-image synthesis.
\newblock In {\em ICML}, 2023.

\bibitem{ogddpm}
Jascha Sohl-Dickstein, Eric Weiss, Niru Maheswaranathan, and Surya Ganguli.
\newblock Deep unsupervised learning using nonequilibrium thermodynamics.
\newblock In {\em ICML}, 2015.

\bibitem{stiennon2020learning}
Nisan Stiennon, Long Ouyang, Jeffrey Wu, Daniel Ziegler, Ryan Lowe, Chelsea
  Voss, Alec Radford, Dario Amodei, and Paul~F Christiano.
\newblock Learning to summarize with human feedback.
\newblock {\em NeurIPS}, 2020.

\bibitem{sun2023evaclip}
Quan Sun, Yuxin Fang, Ledell Wu, Xinlong Wang, and Yue Cao.
\newblock Eva-clip: Improved training techniques for clip at scale.
\newblock {\em arXiv preprint arXiv:2303.15389}, 2023.

\bibitem{cider}
Ramakrishna Vedantam, C Lawrence~Zitnick, and Devi Parikh.
\newblock Cider: Consensus-based image description evaluation.
\newblock In {\em CVPR}, 2015.

\bibitem{wu2023spacetime}
Qiucheng Wu, Yujian Liu, Handong Zhao, Trung Bui, Zhe Lin, Yang Zhang, and
  Shiyu Chang.
\newblock Harnessing the spatial-temporal attention of diffusion models for
  high-fidelity text-to-image synthesis, 2023.

\bibitem{hps}
Xiaoshi Wu, Keqiang Sun, Feng Zhu, Rui Zhao, and Hongsheng Li.
\newblock Better aligning text-to-image models with human preference.
\newblock {\em arXiv preprint arXiv:2303.14420}, 2023.

\bibitem{xu2023imagereward}
Jiazheng Xu, Xiao Liu, Yuchen Wu, Yuxuan Tong, Qinkai Li, Ming Ding, Jie Tang,
  and Yuxiao Dong.
\newblock Imagereward: Learning and evaluating human preferences for
  text-to-image generation, 2023.

\bibitem{yang2022lang}
Yue Yang, Artemis Panagopoulou, Shenghao Zhou, Daniel Jin, Chris
  Callison-Burch, and Mark Yatskar.
\newblock Language in a bottle: Language model guided concept bottlenecks for
  interpretable image classification.
\newblock {\em arXiv preprint arXiv:2211.11158}, 2022.

\bibitem{parti}
Jiahui Yu, Yuanzhong Xu, Jing~Yu Koh, Thang Luong, Gunjan Baid, Zirui Wang,
  Vijay Vasudevan, Alexander Ku, Yinfei Yang, Burcu~Karagol Ayan, et~al.
\newblock Scaling autoregressive models for content-rich text-to-image
  generation.
\newblock {\em TMLR}, 2022.

\bibitem{vlmbow}
Mert Yuksekgonul, Federico Bianchi, Pratyusha Kalluri, Dan Jurafsky, and James
  Zou.
\newblock When and why vision-language models behave like bag-of-words models,
  and what to do about it?
\newblock In {\em ICLR}, 2023.

\bibitem{stackgan}
Han Zhang, Tao Xu, Hongsheng Li, Shaoting Zhang, Xiaogang Wang, Xiaolei Huang,
  and Dimitris Metaxas.
\newblock Stackgan: Text to photo-realistic image synthesis with stacked
  generative adversarial networks.
\newblock In {\em ICCV}, 2017.

\bibitem{zhang2017stack}
Han Zhang, Tao Xu, Hongsheng Li, Shaoting Zhang, Xiaogang Wang, Xiaolei Huang,
  and Dimitris~N. Metaxas.
\newblock Stackgan: Text to photo-realistic image synthesis with stacked
  generative adversarial networks.
\newblock In {\em Proceedings of the IEEE International Conference on Computer
  Vision (ICCV)}, Oct 2017.

\end{thebibliography}
